\documentclass{article}
\usepackage{graphicx}
\usepackage{hyperref}
\usepackage{cite}
\usepackage{float}
\usepackage{tabularx}
\usepackage{titlesec}
\usepackage{amsmath}
\usepackage{booktabs}
\usepackage{caption}
\usepackage{microtype}
\usepackage[left=2.5cm,right=2.5cm,top=2.5cm,bottom=2.5cm]{geometry}
\usepackage{multicol}
\usepackage{enumitem}
\usepackage{colortbl}
\usepackage{xcolor}
\usepackage{multirow}
\usepackage{tikz}
\usepackage{pgfplots}
\pgfplotsset{compat=1.18}
\usepackage{pgf-pie}
\usepackage{tcolorbox}
\usepackage{array}
\usepackage{soul}
\usepackage{ragged2e}

\tcbuselibrary{skins, breakable}

\newcommand{\highlight}[1]{\colorbox{yellow!50}{#1}}

\definecolor{lightgray}{rgb}{0.9,0.9,0.9}
\definecolor{lightblue}{rgb}{0.85,0.90,1.0}

\usetikzlibrary{shapes.geometric, arrows}
\tikzstyle{startstop} = [rectangle, rounded corners, minimum width=3cm, minimum height=1cm, text centered, draw=black, fill=red!30]
\tikzstyle{process} = [rectangle, minimum width=3cm, minimum height=1cm, text centered, draw=black, fill=orange!30]
\tikzstyle{decision} = [diamond, minimum width=3cm, minimum height=1cm, text centered, draw=black, fill=green!30]
\tikzstyle{arrow} = [thick,->,>=stealth]

\title{Towards Leveraging Large Language Models for Automated Medical Q\&A Evaluation}
\author{
  Jack Krolik\thanks{Northeastern University, \texttt{krolik.j@northeastern.edu}}, 
   Herprit Mahal\thanks{Suki AI, \texttt{hmahal@suki.ai}}, 
  Feroz Ahmad\thanks{Suki AI, \texttt{fahmad@suki.ai}}, 
  Gaurav Trivedi\thanks{Suki AI, \texttt{gtrivedi@suki.ai}},
  Bahador Saket\thanks{Suki AI, \texttt{bsaket@suki.ai}}
}
\begin{document}

\maketitle

\begin{abstract}
\fontsize{11pt}{14pt}\selectfont
This paper explores the potential of using Large Language Models (LLMs) to automate the evaluation of responses in medical Question and Answer (Q\&A) systems, a crucial form of Natural Language Processing. Traditionally, human evaluation has been indispensable for assessing the quality of these responses. However, manual evaluation by medical professionals is time-consuming and costly. Our study examines whether LLMs can reliably replicate human evaluations by using questions derived from patient data, thereby saving valuable time for medical experts. While the findings suggest promising results, further research is needed to address more specific or complex questions that were beyond the scope of this initial investigation.
\end{abstract}

\section{Introduction}
Natural Language Processing (NLP) has become a cornerstone in the development of intelligent systems capable of understanding and generating human language. NLP plays a crucial role in extracting insights from vast amounts of unstructured data, enabling numerous applications across various domains. For example, in agriculture, NLP is used to analyze crop health data, interpret satellite imagery, and provide farmers with actionable insights based on weather forecasts, soil conditions, and disease reports \cite{soffosdotai}. In the cybersecurity industry, NLP is employed to analyze and classify threat data, identify malicious activities, and detect phishing attempts, thereby enhancing the ability to respond to and mitigate security risks \cite{bluegoatcyber}. The importance of NLP tasks lies in their ability to bridge the communication gap between humans and machines, making interactions more intuitive and efficient \cite{jurafsky2000speech, manning1999foundations}.

NLP encompasses a variety of tasks, including sentiment analysis \cite{pang2008opinion}, summarization \cite{nenkova2011automatic}, named entity recognition (NER) \cite{nadeau2007survey}, and question answering (Q\&A) \cite{rusu2007open}. In the medical domain, the Q\&A task is extensively used to facilitate clinicians' access to patient information through natural language queries across structured and unstructured data within Electronic Health Records (EHR). Specifically, these Q\&A systems are designed to provide clinicians with accurate and relevant answers to their queries \cite{roberts2021common, friedman2004natural}, and offer advanced search functionalities for easier navigation \cite{campbell2007survey}. These efforts aim to address a major challenge for healthcare professionals, i.e., saving the time and effort required to retrieve patient information from the vast amounts of data stored in medical records \cite{downing2018physician}.

Traditionally, the evaluation of medical Q\&A systems has relied on using manual processes. Medical professionals assess the system responses based on various metrics such as precision, recall, medical correctness, and relevance \cite{demner2009smoking}. While thorough, this manual evaluation is labor-intensive, expensive, and subject to variability among evaluators which can affect the consistency and reproducibility of the results \cite{goodman2017reproducibility}.

Given these challenges, this paper aims to investigate the potential of Large Language Models (LLMs) to automate the evaluation of medical Q\&A systems. LLMs, such as GPT-4o, have demonstrated exceptional performance in generating human-like text and understanding complex queries \cite{brown2020language}. Our research examines if these models can accurately and reliably replicate medical professionals' evaluation processes.

By leveraging the advanced capabilities of LLMs, we seek to reduce the time and cost associated with manual evaluations, allowing medical experts to focus on more sophisticated tasks. We will assess the feasibility of using LLMs to evaluate system responses based on metrics such as relevance, succinctness, medical correctness, hallucination, completeness, and coherence \cite{chiang2023human}. Through this approach, we aim to provide a complementary tool to human evaluation, enhancing efficiency and reliability in the medical domain.

\begin{table}[H]
    \centering
    \small
    \begin{tabularx}{\textwidth}{|>{\raggedright\arraybackslash}X|>{\raggedright\arraybackslash}X|}
        \hline
        \textbf{Question} & \textbf{Ground Truth} \\
        \hline
        How many times has the patient been previously diagnosed with polychondritis?&
        Twice – once on 2022-09-17 and once on 2020-08-14. \\
        \hline
        Is there any risk from the lung nodule? &
        Low risk of lung cancer. Based on a CT scan done at Hopkins after a recommendation from Dr.~Park (cardiologist), the 3 mm lung nodule appears to be an incidental finding. \\
        \hline
        What was the last WBC reading? & 6.2 thou/cumm \\
        \hline
        Has this patient been prescribed treatment for asthma? & Yes, prescribed medications include:
        \begin{itemize}[noitemsep,nolistsep,leftmargin=*]
            \item Albuterol sulfate HFA 90 mcg/actuation aerosol inhaler
            \item Stiolto Respimat 2.5 mcg-2.5 mcg/actuation
            \item ProAir HFA 90 mcg/actuation aerosol inhaler
            \item Ventolin HFA 90 mcg/actuation aerosol
            \item Trelegy Ellipta 100 mcg-62.5 mcg-25 mcg
            \item Methylprednisolone 4 mg tablets (dose pack)
            \item Spiriva Respimat 2.5 mcg/actuation solution for inhalation
            \item Medrol (Pak) 4 mg tablets (dose pack)
        \end{itemize} \\
        \hline
    \end{tabularx}
    \caption{Example questions and corresponding ground truths the medical team developed using patient data.}
    \label{tab:examples}
\end{table}

\section{Dataset for Evaluation}

To evaluate the effectiveness of using LLMs for automating the assessment of responses in medical Question and Answer systems, we collected a comprehensive dataset. This dataset was crucial for ensuring that our evaluation is thorough and representative of real-world scenarios suggested by clinical experts. The dataset includes 94 \emph{Assessment Sets} each of which are comprised of three key components:
\begin{enumerate}
    \item \textbf{Questions}: Medical questions that cover a broad spectrum of medical topics and complexities. These questions were curated to reflect common queries encountered in medical practice.
    \item \textbf{Ground Truth}: A ground truth response, which serves as a benchmark for evaluation. These responses were carefully crafted and validated by medical professionals to ensure accuracy and reliability.
    \item \textbf{Q\&A System Responses}: We sourced responses from our in-house Q\&A system, developed by a team of machine learning engineers, to these questions, a singular yet effective approach to medical NLP.
\end{enumerate}

The collection of this dataset is fundamental for several reasons. \textbf{First}, it allows for a systematic and objective comparison between human evaluations and LLM-based evaluations. Without a well-defined dataset, any assessment would lack the necessary rigor and reproducibility required for academic research \cite{hull1998database}. \textbf{Second}, having a diverse set of questions and responses enabled us to analyze the performance of LLMs across different medical contexts. This is essential for understanding the strengths and limitations of LLMs in providing reliable and contextually appropriate medical advice \cite{williams2020clinical}. \textbf{Finally}, the inclusion of ground truth responses ensures that our evaluations are anchored to a standard of correctness. This is particularly important in the medical domain, where the accuracy of information can have significant implications for patient outcomes \cite{finlayson2021clinician}.

\subsection{Data Anonymization}
In collecting this dataset, we adhered to ethical guidelines to ensure the privacy and confidentiality of any sensitive information. All questions and responses were anonymized, and any identifiable patient information was excluded from the dataset. This adherence to ethical standards is crucial for maintaining the integrity and trustworthiness of our research \cite{mcgraw2013ethical}.

\subsection{Data Collection Methodology}
Two members of our medical operations team, with backgrounds in medical transcription, reviewed the anonymized patient chart data for six patients and subsequently developed a comprehensive set of 94 questions. This list was reviewed by our medical team and verified for accuracy, usefulness, and completeness. These questions covered various patient-related information, including medical history, social history, family history, diagnostic and lab results, as well as operational and administrative aspects. The team then provided corresponding answers, which were carefully crafted to serve as the ground truth for our evaluation.

To ensure the accuracy and relevance of these 94 questions and ground truth responses, a third medical clinician independently reviewed the questions and answers. This review process involved validating the correctness and contextual appropriateness of each answer relative to the corresponding question, using the data available in the database. This multi-layered validation process is critical to maintaining the integrity and reliability of the ground truth dataset. See Table \ref{tab:examples} for examples of the questions and ground truth responses provided by the medical team.

\begin{table}[h]
    \centering
    \renewcommand{\arraystretch}{1.7} % Adjust the value to increase/decrease row spacing
   \footnotesize
    \begin{tabularx}{\textwidth}{|>{\hsize=0.15\hsize}X|>{\hsize=0.43\hsize}X|>{\hsize=0.34\hsize}X|}
        \hline
        \textbf{Metric} & \textbf{Description} & \textbf{Scoring} \\ 
        \hline
        Relevance & Measures the degree to which the response directly addresses the question posed. & 0: Irrelevant, 1: Not relevant, 2: Somewhat relevant, 3: Highly relevant\\  
        Succinctness &  Assesses the conciseness of the response, ensuring that information is communicated efficiently without unnecessary detail. & 0: Not at all, 1: Not so succinct , 2: Mostly succinct, 3: Highly succinct\\
        Medical Correctness & Evaluates the factual and clinical accuracy of the response, which is crucial for patient safety. & 0: Harmful errors, 1: Concerning errors, 2: Benign error, 3: No errors \\ 
        Hallucination & Checks for the presence of any fabricated or inaccurate information not supported by the patient's data.& 0: Harmful hallucinations, 1: Concerning hallucinations, 2: Benign hallucinations, 3: No hallucinations \\
        Completeness & Ensures that the response provides all necessary information required to comprehensively answer the question. & 0: Very incomplete, 1: Somewhat, 2: Mostly incomplete, 3: Very complete \\
        Coherence & Measures the logical flow and clarity of the response, ensuring that it is easily understandable and logically structured. & 0: Inconsistent, 1: Poor coherence, 2: Mostly coherent, 3: Highly coherent\\
        \hline
    \end{tabularx}
    \caption{Evaluation Metrics for LLM Responses in Medical Context. This table presents six key metrics used to assess the quality of responses generated by Large Language Models (LLMs) in medical applications. The ``Metric" column lists the evaluation criteria. The ``Description" column briefly explains each metric's measurement and its importance in a clinical setting. The ``Scoring" column details the scoring system for each metric, ranging from 0 (lowest) to 3 (highest), with specific descriptors for each score level to ensure consistent evaluation across different responses.} \label{tab:evaluation_metrics}
\end{table}

\subsection{Metrics for Response Evaluation}
After collecting the data required for our evaluation, a dedicated group of researchers from our medical team developed a set of rigorous metrics to evaluate the responses generated by the LLMs. This group included two experienced clinicians with several years of practice in both clinical settings and industry. Their extensive experience in evaluating and implementing LLMs for various health applications and systems provided invaluable insights into the development of these metrics. 

The clinicians designed these metrics by reviewing the existing body of work, drawing on their own clinical experience, and past experience evaluating machine learning models across the health industry. These criteria guided the development of a comprehensive and consistent evaluation framework, summarized in Table~\ref{tab:evaluation_metrics}. Relevance and succinctness ensure efficient communication, medical correctness and hallucination prevention are critical for patient safety, while completeness and coherence contribute to the overall usefulness and reliability of the response in a clinical setting. Each metric is associated with specific criteria and a scoring system ranging from 0 (lowest) to 3 (highest), ensuring a thorough assessment of the responses generated by the LLMs.

\subsection{Automating the Evaluation}
\begin{figure}[H]
    \centering
    \includegraphics[width=.6\linewidth]{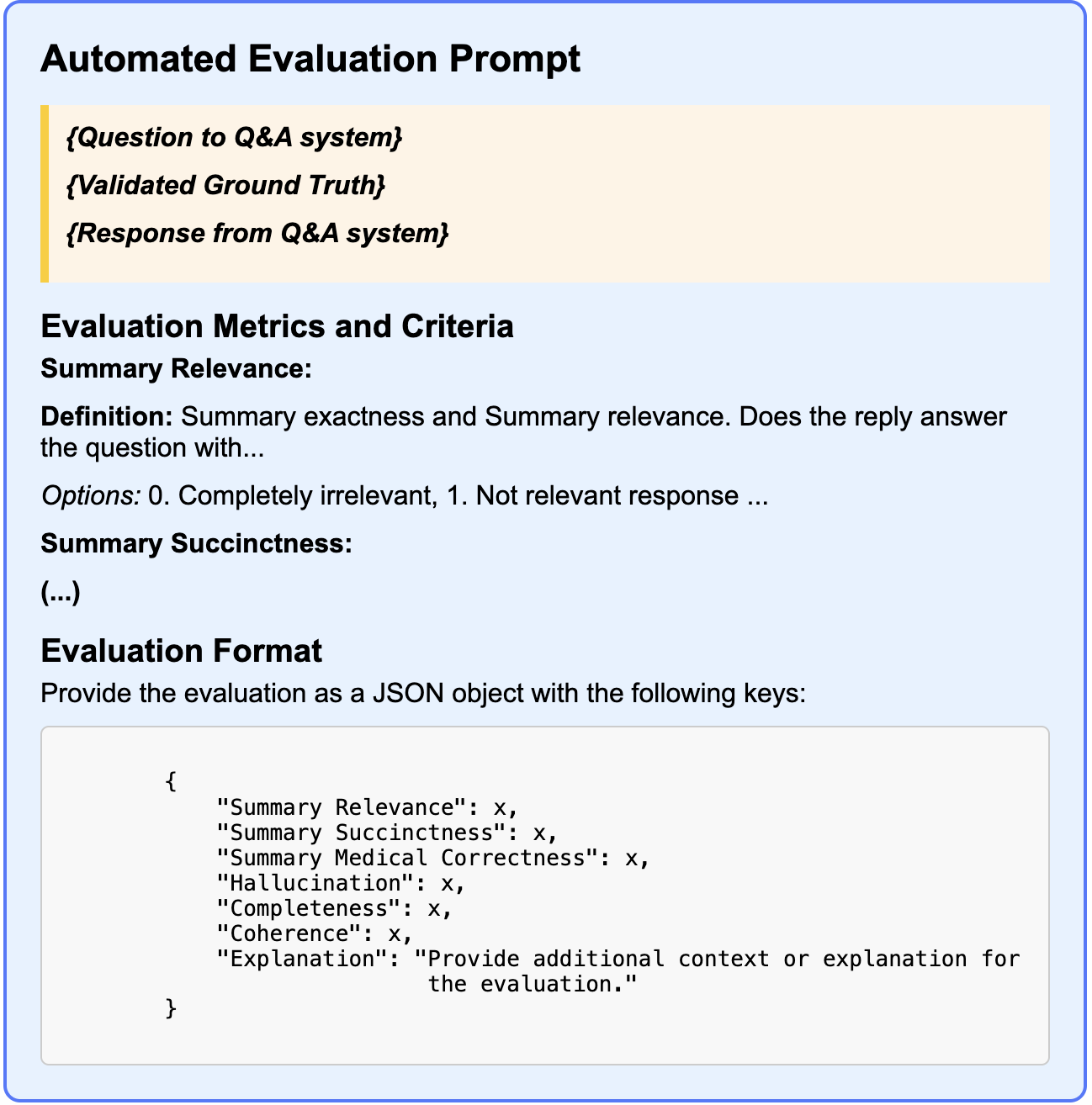}
     \caption{Initial prompt used for automating the evaluation process. It includes:
The image shows an automated evaluation prompt used to guide an LLM in assessing Q\&A system responses. It includes the \textbf{Assessment Set} which is a collection of questions, ground truth answers, and system responses, \textbf{Evaluation Metrics} which are definitions and criteria for the unique metrics to ensure accurate assessments, and the \textbf{Evaluation Format} that covers instructions for structuring responses as a JSON object, enabling consistent integration with existing evaluation processes.}
    \label{fig:initial_prompt}
\end{figure}

We designed a system to automate the evaluation process by identifying a suitable LLM capable of reliably performing the task. We selected ChatGPT-4o for this purpose due to its advanced natural language understanding capabilities, strong performance in medical domain tasks, and ability to follow complex instructions. This made it well-suited for evaluating nuanced medical responses. Our goal was to create a system that could evaluate responses from a Q\&A system in a manner akin to the evaluations conducted by our medical team, but without the labor-intensive manual process. To achieve this, we crafted a prompt to be inputted into ChatGPT-4o (outlined in Figure~\ref{fig:initial_prompt}). 

% \clearpage
The prompt, given to the LLM included the following components:
\begin{itemize}
  \item \textbf{The Assessment Set:} The prompt contained a set of questions, ground truth responses and system responses for each question.
  
  \item \textbf{Definitions, Criteria, and Further Detail regarding each evaluation metric:} This section explained the specific metrics used to evaluate the responses, such as accuracy, relevance, and completeness. By providing detailed definitions and criteria, we ensured the LLM had a comprehensive understanding of what constituted a high-quality response, facilitating more precise and consistent evaluations.
  
  \item \textbf{Response Structure:} The prompt included detailed instructions on how the response should be structured and assessed. This ensured that the LLM could produce an output closely mirroring the format and content of the evaluation spreadsheets used by our medical team.

\end{itemize}
Using ChatGPT-4o we were able to automate the evaluation process, generating a spreadsheet-formatted output similar to the one used by our medical team. This approach streamlined the evaluation process, ensuring consistency, reproducibility, and seamless integration with our existing workflows.

\section{Qualitative Review of Automated Medial Q\&A Evaluation}
A medical team member thoroughly reviewed each LLM evaluation, comparing responses from the LLM to the medical team's answers for each question. The structured evaluation considered patient data, ground truth, and system responses. They then compared the scores assigned by the LLM with those given by the medical team. The evaluation focused on several key aspects:

\begin{itemize}
  \item Highlighted instances where the LLM's performance was on par with or surpassed the medical team’s initial evaluation. This was crucial for identifying the strengths of the LLM in accurately assessing medical responses.
  \item Identified inconsistencies between the LLM’s reasoning, as noted in the "Explanation" section of its evaluation, and the actual scores assigned to each metric. This step was essential for understanding the LLM's decision-making process and pinpointing areas where its logic might be flawed or misaligned with medical standards.
  \item Noted areas where the LLM’s evaluation process could be improved. This included suggestions for refining the LLM's criteria or enhancing its understanding of specific medical concepts.
\end{itemize}

\subsection{Prompt Improvement}
We collaborated with the medical team to refine the prompt by addressing inconsistencies and areas for improvement through an iterative process including: \begin{enumerate}
    \renewcommand{\labelenumi}{\textbf{\arabic{enumi})}}
    \item \textbf{ Adding carefully selected examples}
    \item \textbf{ Developing guidelines to help the LLM prioritize essential information in its evaluations}
\end{enumerate}
The results of these improvements are summarized in Table~\ref{tab:combined_MAE_comparison_prompt_versions}, which shows the progression of the Mean Absolute Error (MAE) across different iterations of the prompt. \\

\begin{table}[htbp]
\centering
\small
\begin{tabular}{@{}l r r r@{}}
\toprule
\multirow{2}{*}{\textbf{Metric}} & \multicolumn{3}{c}{\textbf{Prompt Versions}} \\
\cmidrule(l){2-4}
 & \textbf{Initial} & \textbf{+ Examples} & \textbf{+ Guidelines} \\
\midrule
\rowcolor[gray]{.8}
Summary Precision & 1.39 & 0.89 & 0.69 \\
Summary Succinctness & 0.69 & 0.48 & 0.35 \\
\rowcolor[gray]{.8}
Summary Medical Correctness & 1.46 & 1.15 & 0.62 \\
Hallucination & 1.54 & 1.27 & 0.65 \\
\rowcolor[gray]{.8} 
Completeness & 1.19 & 1.00 & 0.90 \\
Coherence & 0.92 & 0.62 & 0.54 \\
\midrule
\rowcolor{lightblue}
\textbf{Overall MAE} & \textbf{1.20} & \textbf{0.94} & \textbf{0.62} \\
\bottomrule
\end{tabular}
\caption{Comparison of Mean Absolute Error (MAE) across different prompt versions and metrics. Lower MAE values indicate better performance. The "+" in column headers indicates cumulative additions to the prompt: "+ Examples" means examples were added to the initial prompt, and "+ Guidelines" means both examples and guidelines were added.}
\label{tab:combined_MAE_comparison_prompt_versions}
\end{table}

\textbf{Adding Examples:} Our first improvement strategy involved providing the LLM with concrete references to guide its evaluations. We selected 10 representative \textit{Assessment Sets} previously scored by the medical team and asked the medical team to add an ''Explanation'' section for each, detailing their scoring rationale. These examples were then incorporated into the LLM's prompt. The updated LLM was tested on 84 new \textit{Assessment Sets} (excluding the 10 examples to avoid overfitting). By including these carefully chosen examples, we aimed to provide the LLM with clear benchmarks and detailed explanations, enhancing its ability to mirror the medical team's evaluation standards. This strategy led to a 21.67\% improvement, as detailed in Table \ref{tab:combined_MAE_comparison_prompt_versions}.

\textbf{Adding Guidelines:} To further improve performance, we developed generalized guidelines based on the medical team's feedback. These guidelines helped the LLM focus on essential medical information, relevant details, and real-world medical priorities, ensuring comprehensive and relevant assessments. Implementing these guidelines led to a 34.04\% and 50\% improvement in performance from the initial and inclusion of examples prompt respectively, as shown in Table \ref{tab:combined_MAE_comparison_prompt_versions}. \\

By systematically addressing these aspects, the review process aimed to provide a detailed and objective analysis of the LLM's performance. This thorough evaluation was essential for identifying both the strengths and weaknesses of the LLM, ultimately guiding further improvements and ensuring that the automated system could reliably replicate the quality of assessments typically conducted by the medical team. 

\section{Discussion and Future Work}
This section discusses the implications of our study's findings on LLMs' significant potential in automating medical Q\&A evaluations, focusing on time efficiency, added value, and areas for future improvement.

\subsection{Time Efficiency and Resource Allocation}
One of the most substantial advantages of using an LLM for automated evaluations is the considerable reduction in time required. Traditional methods of evaluating responses in medical Q\&A systems often involve manual review by medical professionals, which is not only time-consuming but also resource-intensive. By leveraging LLMs, the evaluation process can be significantly expedited, freeing up valuable time for healthcare providers to focus on direct patient care and other critical tasks.

For instance, the manual evaluation of 94 questions by a medical team typically required around six hours. With the implementation of an LLM, this process was reduced to just 35 minutes: 10 minutes to obtain the LLM-generated responses and an additional 25 minutes for a medical professional to review and finalize the evaluations provided by the LLM. This reduced the time required to evaluate the Q\&A system, leading to more accurate and quicker results for medical professionals. These benefits will be immediate for doctors, including enhanced efficiency and a reduced administrative burden, but ultimately they will create better patient outcomes as well.

\subsection{LLM as a Complementary Evaluation Tool}

Our research revealed that LLMs can serve as valuable complementary tools in the evaluation process, offering a second perspective that enhances the overall assessment quality. This complementary role is illustrated by a specific case from our study as shown in Figure~\ref{fig:case_study}.

This case demonstrates the LLM's capability to catch nuanced errors that human evaluators might overlook, particularly when dealing with extensive medication lists or complex medical information. The LLM's evaluation highlights three key advantages:

\begin{itemize}
    \item \textbf{Error Detection:} LLMs can identify subtle mistakes, such as the inclusion of non-asthma medications, enhancing the overall accuracy of the evaluation, and subsequently the Q\&A system. 
    \item \textbf{Consistency:} LLMs maintain a consistent level of attention to detail across all evaluations, mitigating the risk of human fatigue or oversight.
    \item \textbf{Comprehensive Review:} LLMs can systematically check each aspect of the response against predefined criteria, ensuring a thorough evaluation of both correctness and completeness.
\end{itemize}

% \clearpage
\begin{tcolorbox}[
    enhanced,
    title=Case Study: Asthma Treatment Evaluation,
    colback=white,
    colframe=blue!70!black,
    colbacktitle=blue!70!black,
    fonttitle=\bfseries\normalsize\sffamily,
    coltitle=white,
    attach boxed title to top center={yshift=-2mm},
    boxed title style={sharp corners},
    breakable,
    before upper={\parindent15pt},
    width=\textwidth,
    enlarge top by=5mm,
    enlarge bottom by=5mm,
    boxrule=0.5mm,
    arc=2mm,
    left=2mm,
    right=2mm
]
\RaggedRight
\large
\textbf{Question:} "Has the patient been prescribed treatment for asthma?"

\small
\medskip
\textbf{Ground Truth:}
\highlight{Yes, albuterol sulfate HFA 90 mcg/actuation aerosol inhaler}, Stiolto Respimat 2.5 mcg-2.5 mcg/actuation, \highlight{ProAir HFA 90 mcg/actuation aerosol inhaler}, \highlight{Ventolin HFA 90 mcg/actuation aerosol}, \highlight{Trelegy Ellipta 100 mcg-62.5 mcg-25 mcg}, methylprednisolone 4 mg tablets in a dose pack, Spiriva Respimat 2.5 mcg/actuation solution for inhalation, Medrol (Pak) 4 mg tablets in a dose pack.

\smallskip
\textbf{Q\&A System Response:}
\small
\highlight{Yes}, the patient has been prescribed\highlight{albuterol sulfate 2.5 mg/3 ml (0.083\%) solution for} \highlight{nebulization}, \highlight{albuterol sulfate hfa 90 mcg/actuation aerosol inhaler},\highlight{trelegy ellipta 100 mcg-62.5 mcg-25 mcg powder} \highlight{for inhalation}, \textcolor{red}{\textbf{benzonatate 200 mg capsule}}, and \highlight{ventolin hfa 90 mcg/actuation aerosol inhaler}.

\smallskip
\textbf{Evaluations:}
\footnotesize
\begin{tabular}{l|cccccc}
 & Precision & Succinctness & Correctness & Hallucination & Completeness & Coherence \\
\hline
Medical Team & 3 & 3 & 3 & 3 & 3 & 3 \\
LLM & 3 & \textcolor{red}2 & \textcolor{red}{2} & \textcolor{red}{2} & \textcolor{red}{2} & 3 \\
\end{tabular}

\smallskip
\textbf{LLM Explanation:}
\small
The response is highly relevant and succinct, providing a list of medications that are indeed used for treating asthma. However,\highlight{it is not completely medically correct as it includes \textcolor{red}{\textbf{benzonatate}}}, which is\highlight{a non-narcotic oral medication used to suppress coughs and not typically used in asthma treatment}. This affects the medical correctness, introduces a hallucination, and impacts the completeness of the response.

\end{tcolorbox}
\noindent\begin{minipage}{\textwidth}
\captionsetup{hypcap=false}
\captionof{figure}{Case Study: Comparison of actual and LLM-recommended asthma treatments. Correct recommendations are highlighted in yellow, incorrect in red. The medical team gave perfect scores, while the LLM identified specific issues in medical correctness, hallucination, and completeness, showcasing its potential in medical recommendations.}
\label{fig:case_study}
\captionsetup{hypcap=true}
\end{minipage}

\subsection{Potential Areas for Improvement}

While our results are promising, several areas for improvement have been identified to enhance the LLM's performance and reliability in medical Q\&A evaluation:

\subsubsection{Multi-Model Approach}
Leveraging multiple Large Language Models (LLMs) can significantly enhance the robustness of evaluation systems by incorporating diverse capabilities and perspectives. To implement this effectively, several steps should be taken including utilizing diverse LLMs (e.g., ChatGPT-4, Claude, Pi) to exploit their unique strengths, and assign evaluation metrics tailored to each model's expertise (e.g., medical terminology, context handling). \cite{Brown2021, Anthropic2023, Inflection2024, Zhang2023}. An ensemble method should then be developed to combine the outputs of these models, integrating their evaluations into a cohesive assessment \cite{Kumar2024}. Finally, a machine learning algorithm should be employed to dynamically optimize the weighting of each model's contributions, based on historical performance data, to ensure continuous improvement in evaluation accuracy \cite{Smith2023}. This approach could maximizes the strengths of each LLM and provides a more comprehensive and reliable evaluation system and strategy.

\subsubsection{Iterative Prompt Engineering}

To improve the LLM's evaluation quality, we recommend a continuous process of prompt refinement:

\begin{enumerate}
    \item Regular review sessions with the medical team to analyze LLM performance
    \item Identification of common error patterns or misinterpretations
    \item Incorporation of new examples and guidelines into the prompt
    \item Testing of updated prompts on a diverse set of medical Q\&A scenarios
    \item Quantitative analysis of performance improvements after each iteration
\end{enumerate}

\subsection{Study Limitations and Ethical Considerations}

\subsubsection{Limitations of the Current Study}
While our study demonstrates the potential of LLMs in medical Q\&A evaluation, it is essential to acknowledge its limitations:

\begin{itemize}
    \item \textbf{Sample Size:} Our dataset of 94 assessment sets, though substantial, may not capture all possible scenarios in medical Q\&A. A larger and more representative dataset is needed to ensure greater generalizability across a broader range of medical contexts \cite{Smith2023}.
    
    \item \textbf{Data Source:} Relying on a specific medical database may introduce biases, compromising the accuracy and fairness of LLM assessments and potentially skewing results \cite{Johnson2022}.
    
    \item \textbf{LLM Knowledge Limitations:} Current LLMs may struggle with rare medical conditions or recent research developments not included in their training data. This limitation highlights the need for ongoing updates and refinements to address emerging medical knowledge \cite{Chen2020}.

    \item \textbf{ Single Human Reviewer:} Our study compared the evaluations of the medical operations team with those of the automated model using a single human reviewer. This approach may not account for human expert assessment variability, potentially biasing the review process towards a single perspective. A more robust experiment would compare the evaluation to multiple human reviewers, using separate sets for training and evaluation, aiming for consensus-based validation rather than reliance on a single expert’s opinion. \cite{Gwet2014}
    
\end{itemize}

\subsubsection{Addressing Limitations in Future Research}
To address these limitations, future research should focus on several key areas, while recognizing that our current exploration of using LLMs for automated medical Q\&A evaluation is limited to a specific application and set of questions used in our study. Expanding to other medical domains and incorporating a more diverse and extensive range of questions is crucial for establishing broader applicability. This includes exploring different medical specialties and contexts, as well as utilizing larger, more diverse datasets from multiple medical institutions to enhance the robustness and reliability of LLM evaluations \cite{Wang2024}. Additionally, including rare conditions and a wide spectrum of specialties will better equip models to handle various scenarios \cite{Nguyen2023}. Engaging a diverse panel of medical experts from different specialties to evaluate LLM outputs will help mitigate individual biases and provide a more comprehensive assessment of the models' performance. Furthermore, regularly updating LLMs with the latest medical research and guidelines is essential for improving their accuracy and relevance \cite{Brown2021}. Finally, cross-domain validation—testing the models on questions that bridge multiple specialties or extend into related healthcare domains—will ensure their applicability in complex, real-world medical scenarios beyond the scope of our current study.

\subsubsection{Ethical Considerations}
Ethical considerations are paramount when implementing LLMs in medical contexts. It is crucial to view LLMs as tools designed to augment rather than replace the judgment of medical professionals \cite{Jones2023}. Implementation of LLM-based evaluation systems should adhere to several principles: maintaining human oversight throughout the evaluation process, regularly auditing LLM performance and decision-making processes, and ensuring transparency in the use of AI-assisted evaluation tools \cite{Kumar2024}. Moreover, protecting patient privacy and data security must be a priority, alongside continuous education for medical professionals regarding the capabilities and limitations of LLM-based systems \cite{Lee2022}. By addressing these limitations and ethical considerations, we can work towards a more robust, reliable, and responsible implementation of LLMs in medical Q\&A evaluation.

\section{Conclusion}
This study demonstrates that LLMs, when properly tuned with domain-specific examples and guidelines, can effectively automate the evaluation of medical Q\&A systems. Our iterative approach reduced the Mean Absolute Error to 0.62 on a 0-3 scale, indicating a high level of agreement with medical experts. By automating  these evaluations, LLMs can help medical professionals save valuable time and resources, allowing them to focus more on patient care while still maintaining high-quality evaluations. We encourage future work to expand the scope of questions used in this study, exploring a broader range of scenarios and enhancing the robustness of these systems. While this technology has the potential to significantly reduce the time burden on clinicians, it is crucial to view LLMs as complementary tools rather than replacements for human expertise in medical contexts.

\bibliographystyle{plain}
\bibliography{main}
\end{document}